\documentclass{article} 
\usepackage{iclr2017_conference,times}
\usepackage[utf8]{inputenc}
\usepackage[T1]{fontenc}
\usepackage{graphicx}
\usepackage{amsmath}
\usepackage{amssymb}
\usepackage{listings}
\usepackage{hyperref}
\usepackage{booktabs}
\usepackage{url}

\newcommand{\tbf}[1]{\textbf{#1}}
\newcommand{\tb}[1]{\textbf{#1}}
\newcommand{\tit}[1]{\textit{#1}}

\newcommand{\tm}[1]{\texttt{#1}}

\iclrfinalcopy 
\title{nmtpy: A Flexible Toolkit for Advanced \mbox{Neural} Machine Translation Systems}

\author{Ozan Caglayan, Mercedes Garc\'ia-Mart\'inez, Adrien Bardet, Walid Aransa, \\ \bf Fethi Bougares, Lo\"ic Barrault\\
Laboratoire d'Informatique de l'Université du Maine (LIUM) \\
Language and Speech Technology (LST) Team \\
Le Mans, France \\
}

\begin{document}

\maketitle

\begin{abstract}
In this paper, we present \tit{nmtpy}, a flexible Python toolkit based on Theano
for training Neural Machine Translation and other neural sequence-to-sequence
architectures. \tit{nmtpy} decouples the specification of a network
from the training and inference utilities to simplify the addition of a new
architecture and reduce the amount of boilerplate code to be written. \tit{nmtpy}
has been used for LIUM's top-ranked submissions to WMT Multimodal Machine Translation and
News Translation tasks in 2016 and 2017.
\end{abstract}

\section{Overview}
\tit{nmtpy} is a refactored, extended and Python 3 only version of
\tit{dl4mt-tutorial} \footnote{\url{https://github.com/nyu-dl/dl4mt-tutorial}},
a Theano \citep{theano} implementation of attentive Neural Machine Translation (NMT) \citep{bahdanau2014neural}.

The development of \tit{nmtpy} project which has been open-sourced\footnote{\url{https://github.com/lium-lst/nmtpy}} under
MIT license in March 2017, started in March 2016 as an effort to adapt \tit{dl4mt-tutorial} to multimodal translation models.
\tit{nmtpy} has now become a powerful toolkit where
adding a new model is as simple as deriving from an abstract base class to fill in
a set of fundamental methods and (optionally) implementing a custom data iterator.
The training and inference utilities are as model-agnostic as possible allowing
one to use them for different sequence generation networks
such as multimodal NMT and image captioning to name a few.
This flexibility and the rich set of provided architectures (Section ~\ref{sec:arch})
is what differentiates \tit{nmtpy} from Nematus \citep{nematus} another
NMT software derived from \tit{dl4mt-tutorial}.

\section{Workflow}
\begin{figure}[!htbp]
\begin{center}
\centering
  \includegraphics[width=.8\linewidth]{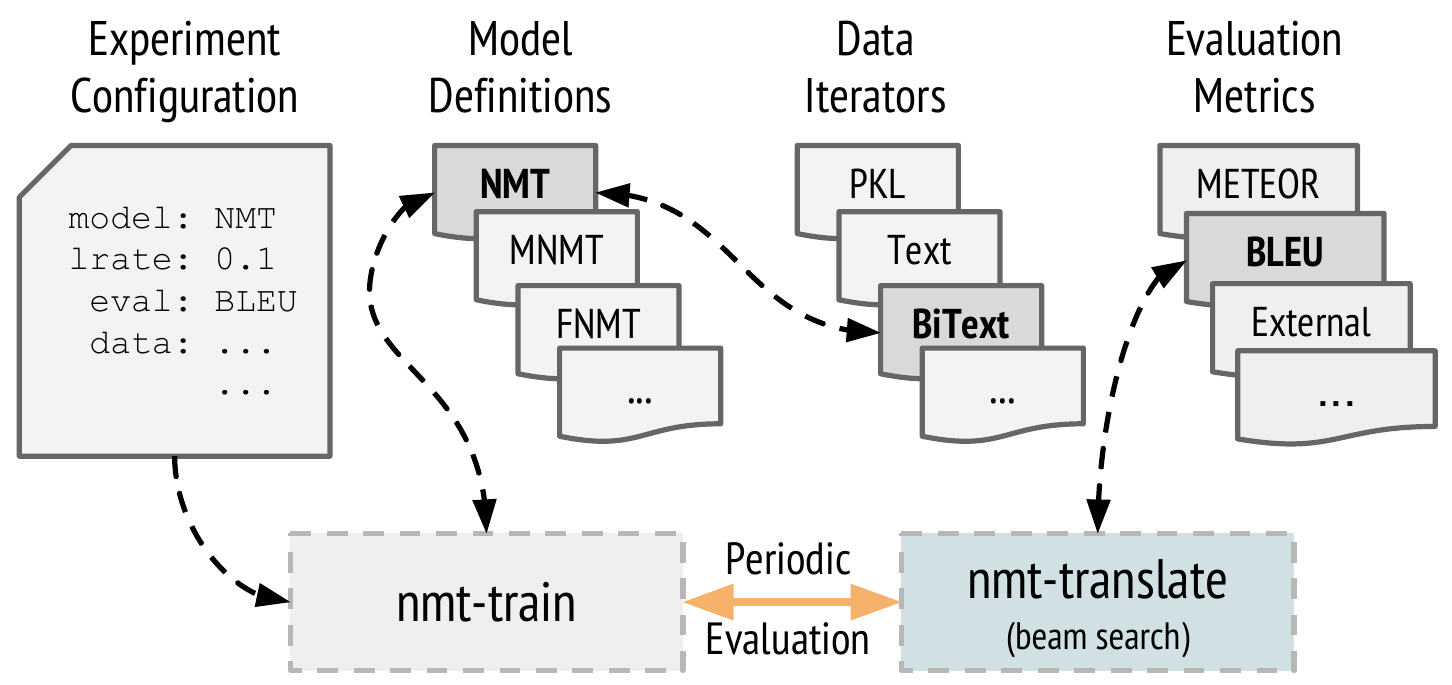}
\end{center}
\caption{The components of \tbf{nmtpy}.}
\label{fig:workflow}
\end{figure}

Figure~\ref{fig:workflow} describes the general workflow of a training session.
An experiment in \tit{nmtpy} is described with a configuration file (Appendix ~\ref{sect:conf}) to ensure
reusability and reproducibility. A training experiment can be simply launched by providing this
configuration file to \tit{nmt-train} which sets up the environment and starts the training.
Specifically \tit{nmt-train} automatically selects a free GPU, sets the seed for all random
number generators and finally creates a model (\tm{model\_type} option) instance.
Architecture-specific steps like data loading, weight initialization and graph construction are delegated to
the model instance. The corresponding log file and model checkpoints are
named in a way to reflect the experiment options determined by the configuration file
(Example: \tm{model\_type-e<embdim>-r<rnndim>-<opt>\_<lrate>...}).

Once everything is ready,
\tit{nmt-train} starts consuming mini-batches of data from the model's iterator to perform
forward/backward passes along with the weight updates.
A validation on held-out corpus is periodically performed to evaluate the generalization
performance of the model. Specifically, after each \tm{valid\_freq} updates, 
\tit{nmt-train} calls the \tit{nmt-translate} utility which will perform 
beam-search decoding, compute the requested metrics and return the results back so that
\tit{nmt-train} can track the progress and save best checkpoints to disk.

Several examples regarding the usage of the utilities are given in Appendix ~\ref{sec:cmdline}.

\subsection{Adding New Architectures}
New architectures can be defined by creating a new file under \tm{nmtpy/models/} using a copy of
an existing architecture and modifying the following predefined methods:
\begin{itemize}
  \item \tm{\_\_init\_\_()}: Instantiates a model. Keyword arguments
    can be used to add options specific to the architecture that will be automatically
    gathered from the configuration file by \tit{nmt-train}.
  \item \tm{init\_params()}: Initializes the layers and weights.
  \item \tm{build()}: Defines the Theano computation graph that will be used during training.
  \item \tm{build\_sampler()}: Defines the Theano computation graph that will be used during
    beam-search. This is generally very similar to \tm{build()} but with sequential RNN steps
    and non-masked tensors.
  \item \tm{load\_valid\_data()}: Loads the validation data for perplexity computation.
  \item \tm{load\_data()}: Loads the training data.
\end{itemize}

\subsection{Building Blocks}
In this section, we introduce the currently available components and features of \tit{nmtpy}
that one can use to design their architecture.

\paragraph{Training}
\tit{nmtpy} provides Theano implementations of stochastic gradient descent (SGD) and
its adaptive variants RMSProp \citep{rmsprop}, Adadelta \citep{adadelta} and Adam \citep{adam}
to optimize the weights of the trained network. A preliminary support for gradient noise \citep{gradientnoise} is available for Adam.
Gradient norm clipping \citep{pascanu2013difficulty} is enabled by default with a threshold of 5 to avoid exploding gradients.
Although the provided architectures all use the cross-entropy objective by their nature,
any arbitrary differentiable objective function can be used since the training loop
is agnostic to the architecture being trained.

\paragraph{Regularization}
A dropout \citep{srivastava2014dropout} layer which can be placed
after any arbitrary feed-forward layer in the architecture is available.
This layer works in inverse mode where the magnitudes are scaled during training instead
of testing. Additionally, L2 regularization loss with a scalar factor defined by \tm{decay\_c} option in the
configuration can be added to the training loss.

\paragraph{Initialization}
The weight initialization is governed by the \tm{weight\_init} option and supports
Xavier \citep{glorotxavier} and He \citep{he2015delving} initialization methods besides orthogonal \citep{saxe2013exact} and random normal.

\paragraph{Layers}
The following layers are available in the latest version of \tit{nmtpy}:
\begin{itemize}
	\item Feed-forward and highway layer \citep{srivastava2015highway}
  \item Gated Recurrent Unit (GRU) \citep{Chung2014}
  \item Conditional GRU (CGRU) \citep{cgru}
	\item Multimodal CGRU \citep{caglayanwmt16,caglayan2016multimodal}
\end{itemize}
Layer normalization \citep{ba2016layer}, a method that adaptively learns to scale and shift
the incoming activations of a neuron, can be enabled for GRU and CGRU blocks.

\paragraph{Iteration}
Parallel and monolingual text iterators with compressed (.gz, .bz2, .xz)
file support are available under the names \tit{TextIterator} and \tit{BiTextIterator}.
Additionally, the multimodal \tit{WMTIterator} allows using image features and
source/target sentences at the same time for multimodal NMT (Section ~\ref{sec:mnmt}).
We recommend using \tm{shuffle\_mode:trglen} when implemented to speed up
the training by efficiently batching same-length sequences.

\paragraph{Post-processing}
All decoded translations will be post-processed if \tm{filter} option is given in the configuration file.
This is useful in the case where one would like to compute automatic metrics on surface forms instead of segmented.
Currently available filters are \tit{bpe} and \tit{compound} for cleaning subword BPE \citep{sennrich2015neural} and
German compound-splitting \citep{sennrich2015joint} respectively.

\paragraph{Metrics}
\tit{nmt-train} performs a patience based early-stopping using either validation perplexity or one of the
following external evaluation metrics:
\begin{itemize}
\item \tm{bleu}: Wrapper around Moses \tit{multi-bleu} BLEU \citep{bleu2002}
\item \tm{bleu\_v13a}: A Python reimplementation of Moses \tit{mteval-v13a.pl} BLEU
\item \tm{meteor}: Wrapper around METEOR \citep{meteor}
\end{itemize}
The above metrics are also available for \tit{nmt-translate} to immediately score the produced hypotheses. Other
metrics can be easily added and made available as early-stopping metrics.

\section{Architectures}
\label{sec:arch}
\subsection{NMT}
The default NMT architecture (\tb{attention}) is based on the original
\tit{dl4mt-tutorial} implementation which differs from \citet{bahdanau2014neural} in the following major aspects:
\begin{itemize}
  \item CGRU decoder which consists of two GRU layers interleaved with attention mechanism.
  \item The hidden state of the decoder is initialized with a non-linear transformation applied to \tit{mean}
        bi-directional encoder state in contrast to \tit{last} bi-directional encoder state.
	\item The Maxout \citep{maxout} hidden layer before the softmax operation is removed.
\end{itemize}

In addition, \tit{nmtpy} offers the following configurable options for this NMT:
\begin{itemize}
  \item \tb{layer\_norm} Enables/disables layer normalization for bi-directional GRU encoder.
  \item \tb{init\_cgru} Allows initializing CGRU with all-zeros instead of mean encoder state.
  \item \tb{n\_enc\_layers} Number of additional unidirectional GRU encoders to stack on top of
    bi-directional encoder.
  \item \tb{tied\_emb} Allows sharing feedback embeddings and output embeddings (2way) or
    all embeddings in the network (3way) \citep{inan2016tying,press2016using}.
  \item \tb{*\_dropout} Dropout probabilities for three dropout layers placed after
    source embeddings (\tm{emb\_dropout}), encoder hidden states (\tm{ctx\_dropout}) and
    pre-softmax activations (\tm{out\_dropout}).
\end{itemize}

\subsection{Factored NMT}
\begin{figure}[!htbp]
\begin{center}
\centering
\includegraphics[width=.5\linewidth]{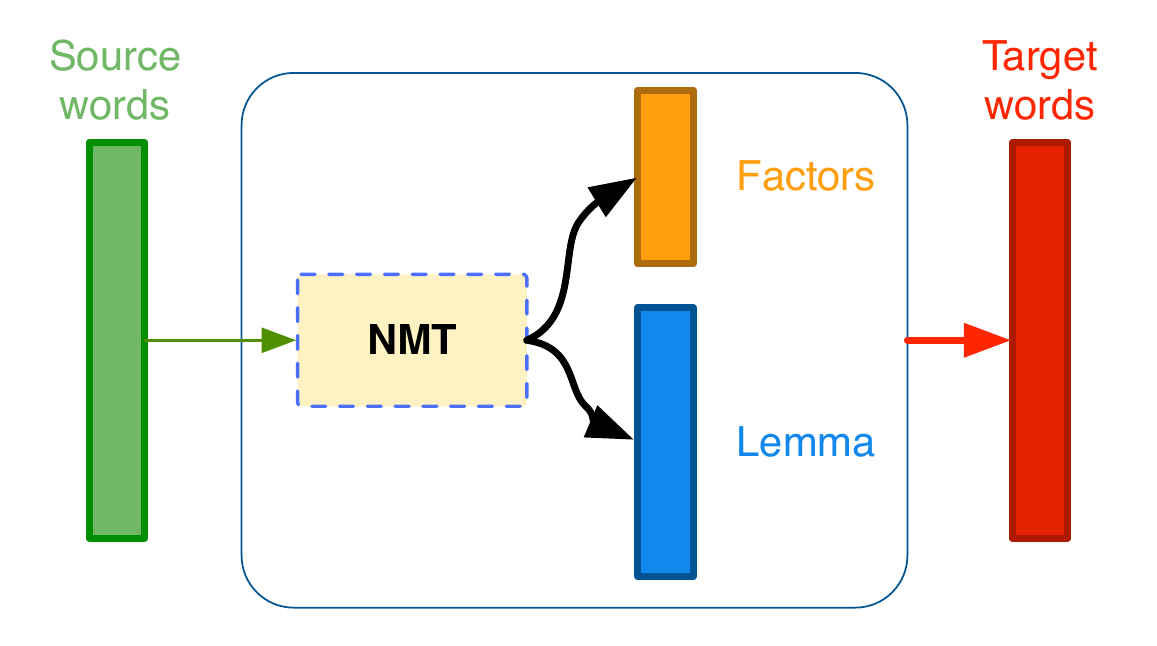}
\end{center}
\caption{Global architecture of the Factored NMT system.}
\label{fig:fnmt}
\end{figure}

Factored NMT (FNMT) is an extension of NMT which is able to generate two output symbols. 
The architecture of such a model is presented in Figure~\ref{fig:fnmt}.
In contrast to multi-task architectures, FNMT outputs share the same recurrence and
output symbols are generated in a synchronous fashion\footnote{FNMT currently uses a dedicated \tit{nmt-translate-factors}
utility though it will probably be merged in the near future.}.

Two FNMT variants which differ in how they handle the output layer are currently available: 
\begin{itemize}
\item \tb{attention\_factors}: the lemma and factor embeddings are concatenated to form a single feedback embedding.
\item \tb{attention\_factors\_seplogits}: the output path for lemmas and factors are kept separate with different
  pre-softmax transformations applied for specialization.
\end{itemize}

FNMT with lemmas and linguistic factors has been successfully used for
IWSLT'16 English$\rightarrow$French \citep{Garcia16iwslt}
and WMT'17\footnote{\url{http://matrix.statmt.org/}} English$\rightarrow$Latvian
and English$\rightarrow$Czech evaluation campaigns.

\subsection{Multimodal NMT \& Captioning}
\label{sec:mnmt}
\begin{figure}[!htbp]
\begin{center}
\centering
   \includegraphics[width=.7\linewidth]{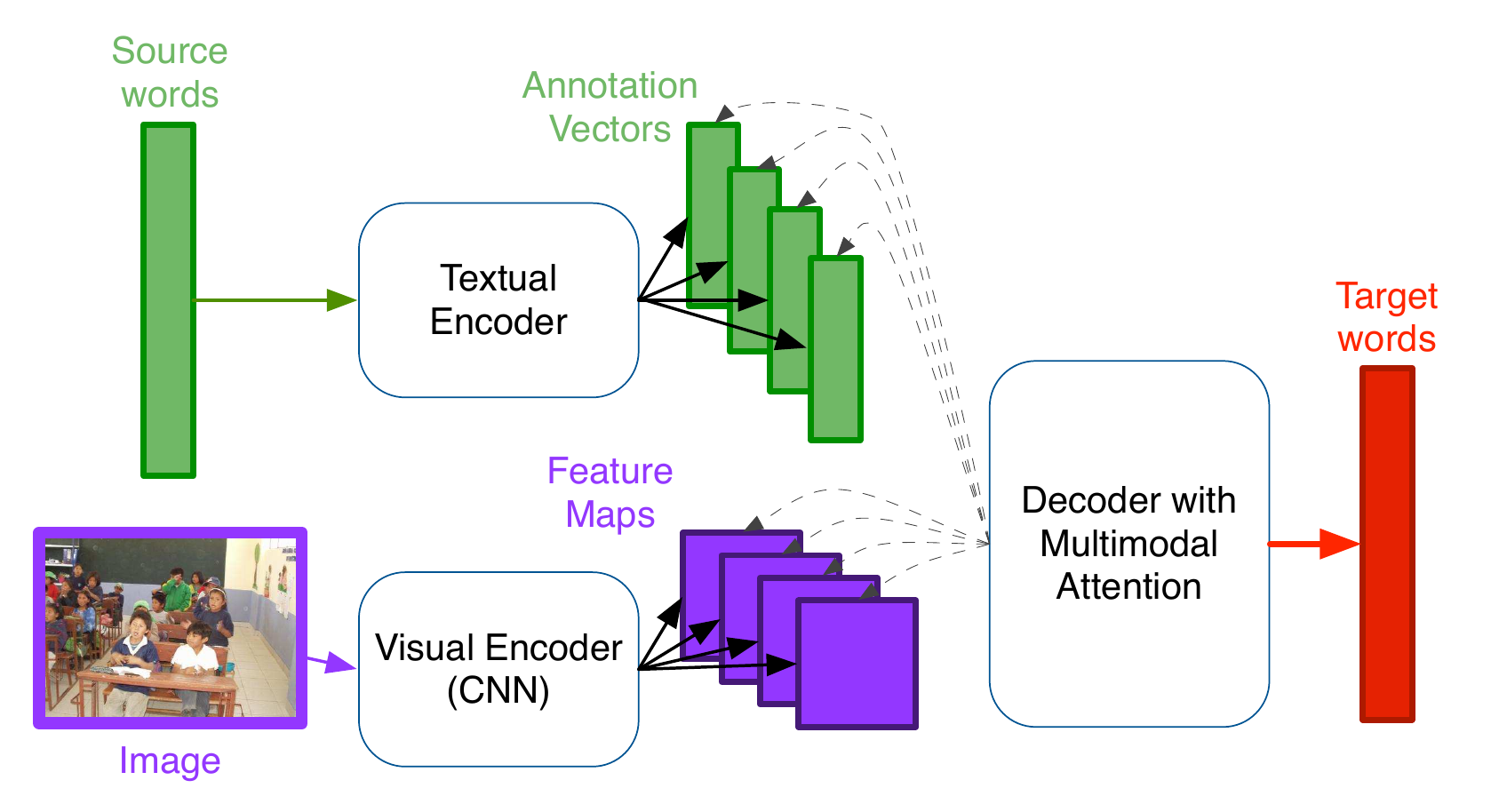}
\end{center}
  \caption{The architecture of multimodal attention \citep{caglayan2016multimodal}.}
\label{fig:fusion}
\end{figure}

We provide several multimodal architectures \citep{caglayanwmt16,caglayan2016multimodal}
where the probability of a target word is conditioned
on source sentence representations and convolutional image features (Figure ~\ref{fig:fusion}).
More specifically, these architectures extends monomodal CGRU into a multimodal one where the attention
mechanism can be shared or separate between input modalities. A late fusion of attended
context vectors are done using either by summing or concatenating the modality-specific
representations.

Our attentive multimodal system for \tit{Multilingual Image Description Generation} track of
\tbf{WMT'16 Multimodal Machine Translation} surpassed the baseline architecture \citep{Elliott2015}
by +1.1 METEOR and +3.4 BLEU and ranked first among multimodal submissions \citep{specia2016shared}.

\subsection{Language Modeling}
A GRU-based language model architecture (\tb{rnnlm}) is available in the repository which can be used
with \tit{nmt-test-lm} to obtain language model scores. 

\subsection{Image Captioning}
A GRU-based reimplementation of \tit{Show, Attend and Tell} architecture \citep{xu2015show} which
learns to generate a natural language description by applying soft attention over 
convolutional image features is available under the name \tb{img2txt}.
This architecture is recently used \footnote{\url{http://www.statmt.org/wmt17/multimodal-task.html}}
as a baseline system for the \tit{Multilingual Image Description Generation} track of
\tbf{WMT'17 Multimodal Machine Translation} shared task.

\section{Tools}
In this section we present translation and rescoring utilities
\tit{nmt-translate} and \tit{nmt-rescore}. Other auxiliary utilities
are briefly described in Appendix~\ref{sec:toollist}.

\subsection{nmt-translate}
\label{nmt-translate}
\tit{nmt-translate} is responsible for translation decoding using the beam-search
method defined by NMT architecture. This default beam-search supports single and ensemble decoding
for both monomodal and multimodal translation models. If a given architecture reimplements the beam-search method
in its class, that one will be used instead.

Since the number of CPUs in a single machine is 2x-4x higher than the number of GPUs and
we mainly reserve the GPUs for training, \tit{nmt-translate} makes use of CPU workers
for maximum efficiency. More specifically, each worker receives a model instance
(or instances when ensembling) and performs the beam-search on samples that
it continuously fetches from a shared queue. This queue is filled by the master process
using the iterator provided by the model.

One thing to note for parallel CPU decoding is that if the \tit{numpy} is linked against a BLAS implementation
with threading support enabled (as in the case with Anaconda \& Intel MKL),
each spawned process attempts to use all available threads in the machine leading to a resource conflict.
In order for \tit{nmt-translate} to benefit correctly from parallelism, the number of threads per process is
thus limited \footnote{This is achieved by setting \tm{X\_NUM\_THREADS=1} environment variable where \tm{X} is one of
\tm{OPENBLAS,OMP,MKL} depending on the \tit{numpy} installation.} to 1. The impact of this setting and the overall
decoding speed in terms of words/sec (wps) are reported in (Table ~\ref{tbl:blas}) for a medium-sized
En$\rightarrow$Tr NMT with $\sim$10M parameters.

\begin{table}[ht]
\centering
\resizebox{.6\linewidth}{!}{%
\begin{tabular}{rrrrr}
\toprule
\tbf{\# BLAS Threads} & \tbf{Tesla K40} & \tbf{4 CPU}  & \tbf{8 CPU} & \tbf{16 CPU} \\ \midrule
Default               & 185 wps         & 26  wps      & 25 wps      & 25 wps       \\
\tb{Set to 1}         & 185 wps         & 109 wps      & 198 wps     & 332 wps      \\ \bottomrule
\end{tabular}}
\caption{Median beam-search speed over 3 runs with beam size 12: decoding on a single Tesla K40 GPU is rougly
equivalent to using 8 CPUs (Intel Xeon E5-2687v3).}
  \label{tbl:blas}
\end{table}

\subsection{nmt-rescore}
A 1-best plain text or $n$-best hypotheses file can be rescored with \tit{nmt-rescore} using either a single
or an ensemble of models. Since rescoring of a given hypothesis
simply means computing the \tit{negative log-likelihood} of it given the source sentence,
\tit{nmt-rescore} uses a single GPU to efficiently compute the scores in batched mode. See Appendix ~\ref{sec:cmdline}
for examples.

%

\section{Conclusion}
We have presented \tit{nmtpy}, an open-source sequence-to-sequence framework
based on \tit{dl4mt-tutorial} and refined in many ways
to ease the task of integrating new architectures. The toolkit
has been internally used in our team for tasks ranging from
monomodal, multimodal and factored NMT to image captioning and
language modeling to help achieving top-ranked submissions during
campaigns like IWSLT and WMT.

\subsubsection*{Acknowledgments}
This work was supported by the French National Research Agency (ANR) through the CHIST-ERA M2CR project, under the contract number ANR-15-CHR2-0006-01\footnote{\texttt{http://m2cr.univ-lemans.fr}}.

\bibliography{paper}
\bibliographystyle{iclr2017_conference}

\newpage
\appendix

\section{Configuration File Example}
\label{sect:conf}

\begin{lstlisting}[frame=single,language=bash]
# Options in this section are consumed by nmt-train
[training]
model_type: attention 	# Model type without .py
patience: 20 		# early-stopping patience
valid_freq: 1000 	# Compute metrics each 1000 updates
valid_metric: meteor 	# Use meteor during validations
valid_start: 2 		# Start validations after 2nd epoch
valid_beam: 3 		# Decode with beam size 3
valid_njobs: 16 	# Use 16 processes for beam-search
valid_save_hyp: True 	# Save validation hypotheses
decay_c: 1e-5 		# L2 regularization factor
clip_c: 5 		# Gradient clip threshold
seed: 1235 		# Seed for numpy and Theano RNG
save_best_n: 2 		# Keep 2 best models on-disk
device_id: auto 	# Pick 1st available GPU
snapshot_freq: 10000 	# Save a resumeable snapshot
max_epochs: 100


# Options below are passed to model instance
[model]
tied_emb: 2way 		# weight-tying mode (False,2way,3way)
layer_norm: True  	# layer norm in GRU encoder
shuffle_mode: trglen 	# Shuffled/length-ordered batches
filter: bpe 		# post-processing filter(s)
n_words_src: 0 		# limit src vocab if > 0
n_words_trg: 0 		# limit trg vocab if > 0
save_path: ~/models 	# Where to store checkpoints
rnn_dim: 100 		# Encoder and decoder RNN dim
embedding_dim: 100 	# All embedding dim
weight_init: xavier
batch_size: 32
optimizer: adam
lrate: 0.0004
emb_dropout: 0.2 	# Set dropout rates
ctx_dropout: 0.4
out_dropout: 0.4


# Dictionary files produced by nmt-build-dict
[model.dicts]
src: ~/data/train.norm.max50.tok.lc.bpe.en.pkl
trg: ~/data/train.norm.max50.tok.lc.bpe.de.pkl


# Training and validation data
[model.data]
train_src     : ~/data/train.norm.max50.tok.lc.bpe.en
train_trg     : ~/data/train.norm.max50.tok.lc.bpe.de
valid_src     : ~/data/val.norm.tok.lc.bpe.en
valid_trg     : ~/data/val.norm.tok.lc.bpe.de
valid_trg_orig: ~/data/val.norm.tok.lc.de
\end{lstlisting}

\hfill 

\section{Usage Examples}
\label{sec:cmdline}
\begin{lstlisting}[frame=single,language=bash,caption=Example usage patterns for \tit{nmt-train}.,captionpos=b]
# Launch an experiment
$ nmt-train -c wmt-en-de.conf

# Launch an experiment with different architecture
$ nmt-train -c wmt-en-de.conf 'model_type:my_amazing_nmt'

# Change dimensions
$ nmt-train -c wmt-en-de.conf 'rnn_dim:500' 'embedding_dim:300'

# Force specific GPU device
$ nmt-train -c wmt-en-de.conf 'device_id:gpu5'
\end{lstlisting}

\begin{lstlisting}[frame=single,language=bash,caption=Example usage patterns for \tit{nmt-translate}.,captionpos=b]
# Decode on 30 CPUs with beam size 10, compute BLEU/METEOR
# Language for METEOR is set through source file suffix (.en)
$ nmt-translate -j 30 -m best_model.npz -S val.tok.bpe.en \
                -R val.tok.de -o out.tok.de -M bleu meteor -b 10

# Generate n-best list with an ensemble of checkpoints
$ nmt-translate -m model*npz -S val.tok.de \
                -o out.tok.50best.de -b 50 -N 50

# Generate json file with alignment weights (-e)
$ nmt-translate -m best_model.npz -S val.tok.bpe.en \
                -R val.tok.de -o out.tok.de -e
\end{lstlisting}

\begin{lstlisting}[frame=single,language=bash,caption=Example usage patterns for \tit{nmt-rescore}.,captionpos=b]
# Rescore 50-best list with ensemble of models
$ nmt-rescore -m model*npz -s val.tok.bpe.en \
                -t out.tok.50best.de \
                -o out.tok.50best.rescored.de
\end{lstlisting}

\section{Description of the provided tools}
\label{sec:toollist}

\paragraph{nmt-build-dict}
Generates \tit{.pkl} vocabulary files from preprocessed corpus. A single/combined
vocabulary for two or more languages can be created with \tm{-s} flag.

\paragraph{nmt-extract}
Extracts arbitrary weights from a model snapshot which can further be used
as pre-trained weights of a new experiment or analyzed using visualization techniques
(especially for embeddings).

\paragraph{nmt-coco-metrics}
A stand-alone utility which computes multi-reference BLEU, METEOR, CIDE-r \citep{vedantam2015cider} and
ROUGE-L \citep{rouge2004} using MSCOCO evaluation tools \citep{chen2015microsoft}. Multiple systems
can be given with \tm{-s} flag to produce a table of scores.

\paragraph{nmt-bpe-(learn,apply)}
Copy of subword utilities \footnote{https://github.com/rsennrich/subword-nmt} \citep{sennrich2015neural}
which are used to first \tit{learn} a BPE segmentation model over a given corpus file and then \tit{apply} it to new sentences.

\paragraph{nmt-test-lm}
Computes language model perplexity of a given corpus.

\end{document}